\documentclass[]{spie}  

\usepackage{amsmath,amsfonts,amssymb}
\usepackage{graphicx}
\usepackage[colorlinks=true, allcolors=blue]{hyperref}

\usepackage{lipsum}
\usepackage{xcolor}
\usepackage{bm}
\usepackage{mathtools}

\title{Orientation stabilization in a bioinspired bat-robot using integrated mechanical intelligence and control}

\author{Eric Sihite}
\author{Andrew Lessieur}
\author{Pravin Dangol}
\author{Akshath Singhal}
\author{Alireza Ramezani}
\affil{SiliconSynapse Laboratory at Northeastern University, Boston, USA}

\authorinfo{Further author information: (Send correspondence to Alireza Ramezani)\\  Alireza Ramezani: E-mail: a.ramezani@northeastern.edu, Telephone: 1 (734) 604 1214}

\begin{document} 
\maketitle

\begin{abstract}

Our goal in this work is to expand the theory and practice of robot locomotion by addressing critical challenges associated with the robotic biomimicry of bat aerial locomotion. Bats wings exhibit fast wing articulation and can mobilize as many as 40 joints within a single wingbeat. Mimicking bat flight can be a significant ordeal and the current design paradigms have failed as they assume only closed-loop feedback roles through sensors and conventional actuators while ignoring the computational role carried by morphology. In this paper, we propose a design framework called \textit{Morphing via Integrated Mechanical Intelligence and Control (MIMIC)} which integrates a small and low energy actuators to control the robot through a change in morphology. In this paper, using the dynamic model of Northeastern University's \textit{Aerobat}, which is designed to test the effectiveness of the MIMIC framework, it will be shown that computational structures and closed-loop feedback can be successfully used to mimic bats stable flight apparatus. 

\end{abstract}


\section{INTRODUCTION}
\label{sec:intro}  


In recent years, there has been increased focus on making our residential spaces smarter, safer, more efficient and closer to the materialization of the concept of smart cities \cite{lee_trends_2014}. As a result, safety and security robots are gaining ever growing importance \cite{pavlidis_urban_2001} and drive a lucrative market. Smart cities market was valued at USD 624.81 billion in 2019 and is expected to reach USD 1712.83 billion by 2025 \cite{noauthor_smart_nodate}. This increases the demand for a highly mobile and energy efficient drone that is robust, fail-safe, and can operate in a wide range of operating conditions which sometimes can be hazardous for the robot.

The drones that are commercially available for fulfilling this demand can be broadly categorized into ground and aerial robots. Ground robots are among the most used as security robots despite their well known limitations, such as relatively low mobility, difficulty in negotiating rough terrain, difficulty in reaching vantage positions for surveillance, limited operation time, and the potential of colliding with humans in crowded spaces \cite{tiddi_robotcity_2020,koolen_design_2016,fahmi_passive_2019,dario_bellicoso_perception-less_2016,bretl_testing_2008,pardo_evaluating_2016,mastalli_trajectory_2017,mastalli_-line_2015,winkler_gait_2018,aceituno-cabezas_simultaneous_2018,dai_planning_2016}. 

Aerial robots, on the other hand, possess superior mobility and is extremely suitable for civic surveillance and monitoring applications. However, despite these advantages, Micro Aerial Vehicles (MAVs) have a very limited contribution in this market. Current state-of-the-art MAVs with fast rotating propellers and rigid structures pose extreme dangers to humans, e.g., they can cause penetrating injury resulting in blood loss and massive destruction of the human body \cite{arterburn_david_faa_2017}. Furthermore, they are unable to survive crashes which could be unavoidable in unstructured environments and have a highly limiting operation time, typically in the range of thirty to forty minutes. That said, the application of safety features in these systems have not solved the problems and their operations in residential spaces have remained limited by strict rules from the Federal Aviation Administration (FAA) \cite{arterburn_david_faa_2017}. To be able to use MAVs in civic applications, there is a need to transform their safety and efficiency.

\begin{figure*}[t]
    \centering
    \includegraphics[width=0.99\linewidth]{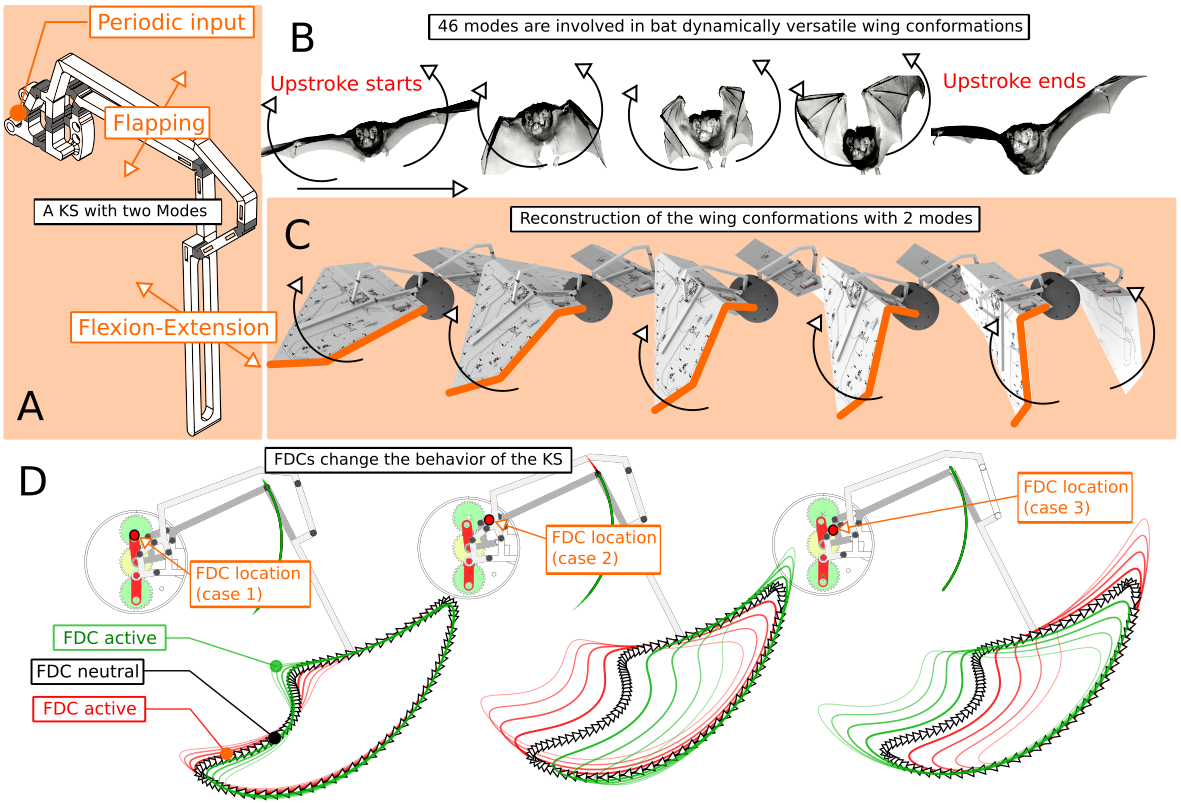}
    \caption{The kinetic sculpture (KS) design and the motivation to mimic a bat's natural flapping gait. (A) Illustrates the KS made of monolithically fabricated rigid and flexible materials. (B) Depicts bat flapping gait with the wing folding/expansion within one flapping cycle. (C) Illustrates simulated Aerobat wingbeat cycle. (D) Shows the sensitivity analysis resutls. Feedback-Driven Components (FDCs) used in the KS can change the behavior of the structure which can be leveraged for flight design purpose.}
    \label{fig:justification}
\end{figure*}

The research goal in this work is to expand the theory and practice of robot locomotion by addressing critical challenges associated with the robotic biomimicry of bat aerial locomotion. The resulting MAVs with morphing bodies will be safe, agile and energy efficient owing to their articulated, soft wings and will autonomously operate with long operation lifespans. Bat membranous wings possess unique functions \cite{tanaka_flexible_2015} that make them a good example to take inspiration from and transform safety, agility and efficiency of current aerial drones. In contrast with other flying vertebrates, bats have an extremely articulated musculoskeletal system (Fig.~\ref{fig:justification}-B) which is key to their body impact survivability and their impressively adaptive and multimodal locomotion behavior \cite{riskin_quantifying_2008}. Bats exclusively use this capability with their structural flexibility to generate the controlled force distribution on each membrane wing. Wing flexibility, complex wing kinematics, and fast muscle actuation allow these creatures to change their body configuration within a few tens of milliseconds. These characteristics are crucial to their unrivaled agility and energetic efficiency \cite{azuma_akira_biokinetics_2006}.

Complex locomotion styles achieved through such synchronous movements of many joints are showcased by several other species. There is an urgency for new paradigms providing insight on how these animals with small brains and muscles, which has limited computational and actuation power, mobilize and regulate so many of their joints. The widely used paradigms have failed to copy bat flight \cite{bahlman_design_2013,colorado_biomechanics_2012} because they assume only closed-loop feedback roles and ignore computational roles carried out by morphology. To respond to the urgency, we propose a framework called Morphing via Integrated Mechanical Intelligence and Control (MIMIC). In this work, using simulation results, it will be shown that the MIMIC framework can be successfully used to mimic bats flight apparatus known for their pronounced, fast wing articulations, e.g., bats can mobilize as many as forty joints during a single wingbeat, with some joints reaching to over one thousand degrees per second in angular speed.

We expand upon our previous work with the bat robots \cite{ramezani_biomimetic_2017, hoff_synergistic_2016, hoff_optimizing_2018, hoff_reducing_2017, ramezani_towards_2020} where we will apply the MIMIC framework to our most recent morphing wing design in [\citenum{sihite_computational_2020}]. This design captures the elbow flexion and extension that allow the wing to fold during upstroke. The MIMIC framework will be the continuation of our past attempts at developing control framework for bio-inspired flapping wing drones \cite{ramezani_lagrangian_2015, hoff_optimizing_2018, hoff_trajectory_2019,ramezani_alireza_nonlinear_2016,ramezani_describing_2017,syed_rousettus_2017,sihite_enforcing_2020}.

\section{ROBOT AND CONTROL FRAMEWORK OVERVIEW}

\begin{figure}[t]
    \centering
    \includegraphics[width=0.6\linewidth]{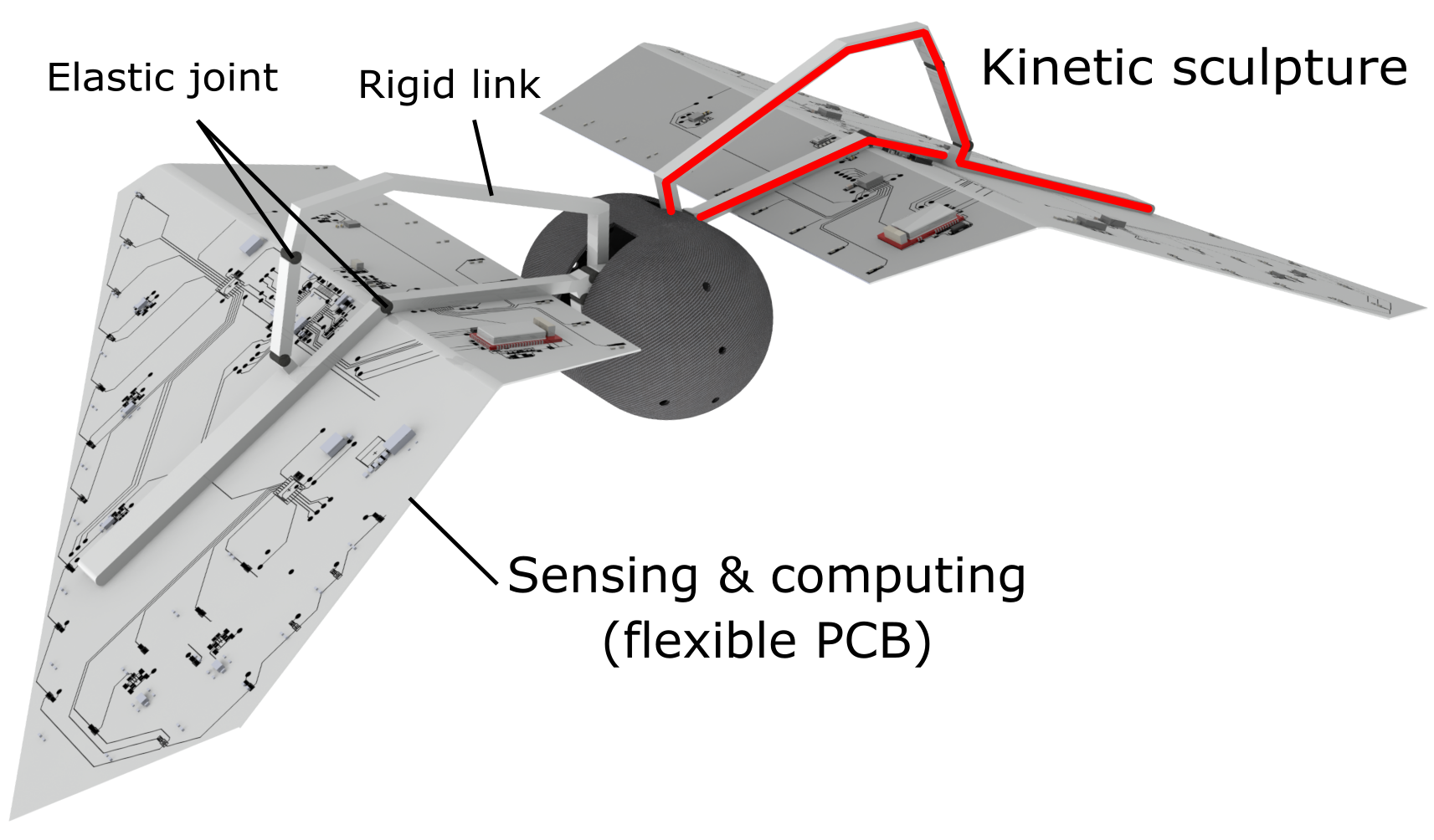}
    \caption{Illustration of Northeastern University's \textit{Aerobat}.}
    \label{fig:cover_photo}
\end{figure}

\begin{figure}[t]
\centering
\includegraphics[width=0.5\linewidth]{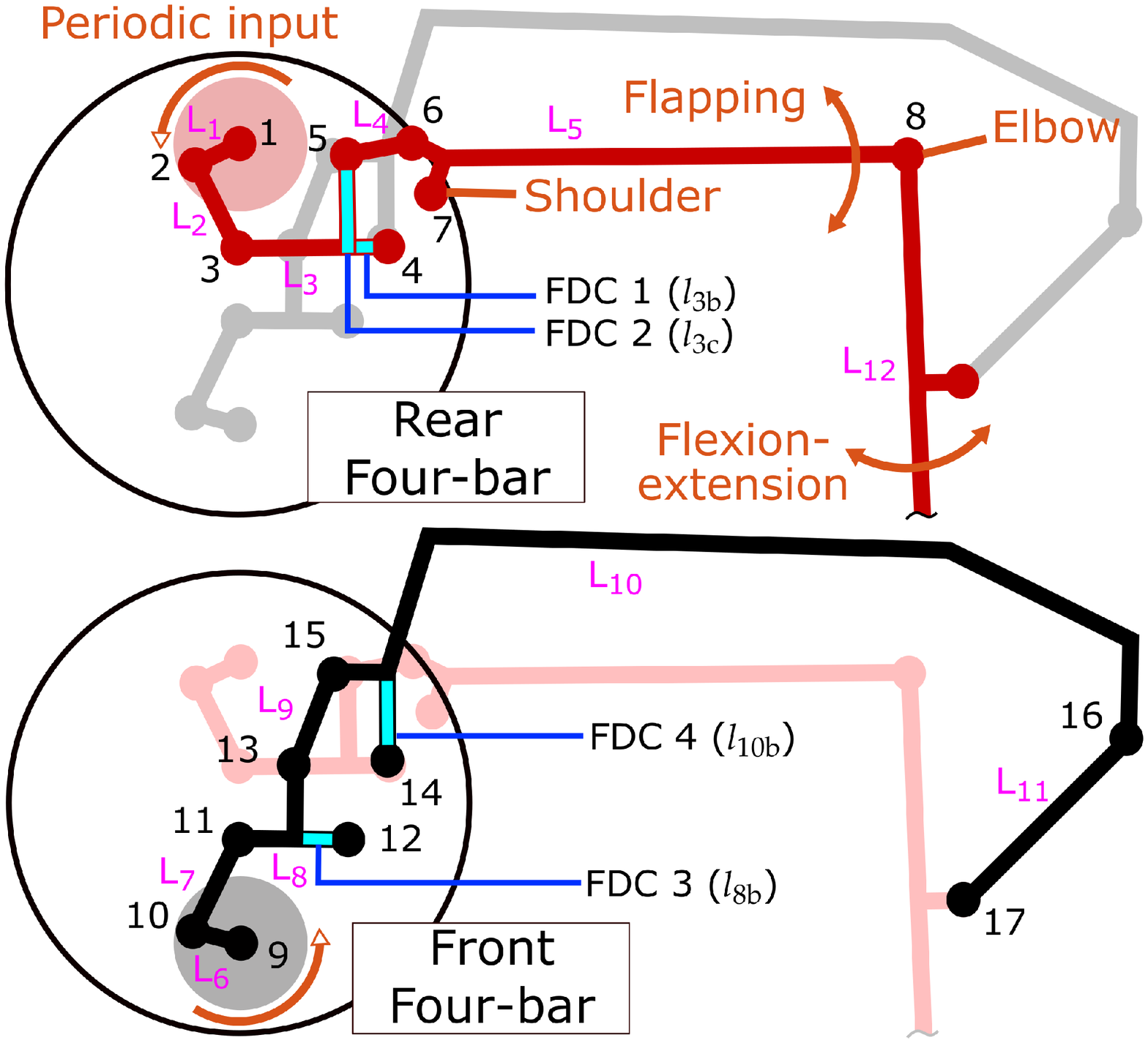}
\caption{Shows the wing structure which is composed of a network of mechanical linkages. The system consists of 12 links ($\{L_1, \dots, L_{12}\}$) and 17 joints/hinges ($\{j_1, ...,j_{17} \}$). Joints 1 and 9 are the gears' center of rotation where they will be driven up to 10 Hz in angular speed. Four FDCs adjust the length of the linkage to change the resulting end-effector trajectories and facilitate control (see the sensitivity analysis results shown above.} 
\label{fig:wing_structure}
\end{figure}

The Northeastern University's \textit{Aerobat}, shown in Fig.~\ref{fig:cover_photo}, is a flapping wing robot with a maximum wingspan of 30 cm. This robot is designed for studying flapping wing flight inspired from bats by incorporating dynamic wing morphing into the armwing design. This robot utilizes a computational structure, which we refer using the term \textit{kinetic sculpture} (KS), fabricated monolithically using both rigid and flexible materials \cite{sihite_computational_2020}. The robot design and the FDC design ideas will be discussed below.

\subsection{Robot and Kinetic Sculpture Design}

The KS, shown in Fig.~\ref{fig:justification}-A captures two of the important modes in a bat flapping gait, i.e., the plunging and elbow flexion/extension shown in Fig.~\ref{fig:justification}-B. These two modes form the dynamic wing morphing shown in bats flight where the wing folds and expands during the upstroke and downstroke, respectively. The wing folding minimize negative lift generated in addition to reducing wing inertia which allow the wing to translate faster during the upstroke. This morphing capability results in a highly efficient flapping gait that we sought in our design. 

The KS' linkage mechanism, shown in Fig.~\ref{fig:wing_structure}, is composed of 12 linkages and 17 joints per wing. The KS of both wings are synchronously actuated by a single motor using a series of spur gears, cranks, and four-bar mechanism which is embedded in the KS linkage design. We implemented the KS into our design to achieve the wing folding during the upstroke as shown in Fig.~\ref{fig:justification}-C. 

Our robot is very likely to be pitch unstable, which is expected in a tail-less ornithopters, and we require an active or passive stabilizer to regulate the robot's attitude. We achieve active stabilization through \textit{Mechanical Intelligence}, which we will discuss this in more detail below.

\subsection{Control Framework Through Morphological Computation}

A more sophisticated structure is proposed to add control and morphological freedom into this design. We propose to use Feedback-Driven Components (FDC), a small and low-power actuator, to adjust the length of some links as shown in Fig. \ref{fig:wing_structure}. The change in armwing morphology results in a change in the end effector trajectories, as shown in Fig.~\ref{fig:justification}-D. This change affects the aerodynamic forces generated by the wing, therefore we can facilitate control through a change body morphology using simple and low power actuation components. This framework is relevant to the concept of \textit{Mechanical Intelligence}, where a simple control action can influence and achieve complex manipulation in the body morphology which is commonly seen in nature \cite{hauser_role_2012}. We refer to this control framework as \textit{Morphing via Integrated Mechanical Intelligence and Control (MIMIC)} and we will show the feasibility of closed loop feedback using this framework through simulation in the next Section \ref{sec:simulation}.

The KS conformation parameters have a varying degree of sensitivity on how they affect the resulting flapping trajectory. We have analyzed the sensitivity of our KS design parameters which is outlined in [\citenum{sihite_computational_2020}]. The linkage parameters closer to the crank arm is more sensitive than the distal linkages, therefore they are good candidates for our FDC placements. Furthermore, a change in linkage conformation which actuates the radius linkages will not affect the humerus linkage trajectory, allowing us to change these trajectories independently from each other. We selected linkage parameters $l_{3b}$, $l_{3c}$, $l_{8b}$, and $l_{10b}$ for our FDC placements due to their high sensitivity, as shown in Fig.~\ref{fig:wing_structure}. The modeling and control approach will be discussed in more detail in the next section.

\section{SYSTEM MODELING AND SIMULATION}
\label{sec:simulation}

\begin{figure}[t]
\centering
\includegraphics[width=0.5\linewidth]{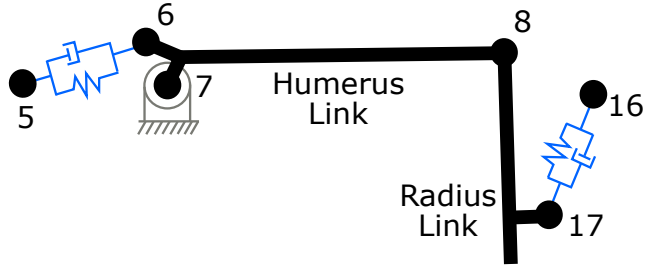}
\caption{Shows the wing massed subsystem which is composed of the humerus and radius links. The external forces and torques imposed by the massless kinematic chain and flexible hinges act on this subsystem.} 
\label{fig:massed_subsystem}
\end{figure}

\subsection{Dynamic Modeling}

The flexible joints of the KS can be modeled using a combination of linear and rotational spring as outlined in \citenum{vogtmann_characterization_2013}. However, as this can be very computationally expensive, we simplified the system into massed and massless subsystems. The massed subsystem is composed of the the body and both wings' humerus and radius linkages where the wing membranes are attached. In this subsystem, the joints are modeled using linear spring and damper, as shown in Fig.~\ref{fig:massed_subsystem}. The remaining linkages are assumed to be massless and modeled using rigid kinematic relationship. The position and velocity of the massless joints 5 and 16 will be used to drive the massed linkage components.

The massless system mode can be derived from kinematic constraints, derived as follows:
\begin{equation}
    \bm q_1 =[\theta_1,\theta_2, \theta_4,\theta_9,\theta_{10},\theta_{12}, \theta_{13},\theta_{14}, l_{3b}, l_{3c}, l_{8b}, l_{10b}]^\top \in \mathbb{R}^{12}
    \label{eq:kinematic_states}
\end{equation}
\begin{equation}
    \bm{u}_k = [u_g, u_{3b}, u_{3c}, u_{8b}, u_{10b}]^\top \in \mathbb{R}^{5}
    \label{eq:kinematic_inputs}
\end{equation}
\begin{equation}
    M_{k}(\bm{q}_k) \ddot{\bm{q}}_k+ \bm{h}_{k} (\bm{q}_k,\dot{\bm{q}}_k) = B_k\,\bm{u}_k,
    \label{eq:kinematic_eom}
\end{equation}
where $\bm q_k$ is the states of the massless subsystems, and $\bm u_k$ is the input to the kinematic constraints. $\bm q_k$ is composed of the joint angles ($\theta$) and FDC lengths states ($l$) while $\bm u_k$ is composed of the accelerations of the crank ($u_g$) and four FDC's length adjustments ($u_{3b}$, $u_{3c}$, $u_{8b}$, $u_{10b}$). The positions of joints 5 and 16 can be derived from $\bm q_k$ which will be used in the dynamical subsystem to drive the humerus and radius links.

The dynamical subsystem can be derived using Euler-Lagrangian equation of motion, which follows a similar derivation to our previous work in [\citenum{sihite_computational_2020}]. We also derived the body's attitude using the modified Lagrangian following Hamilton's principle where the rotation matrix is used to define the Lagrangian. The dynamical equation of motion have 10 DOF which can be derived using the standard form:
\begin{equation}
\begin{gathered}
    M_d(\bm q_d, R_B) [\ddot{\bm{ q}}_d^\top, \dot{\bm \omega}^{B\,\top}]^\top + \bm h_d (\bm q_d, \dot{\bm q}_d, R_b, \bm \omega^B) = B_s\bm u_s + \sum_{k=1}^{N_s} B_{a,k} \bm u_{a,k}, \\
    \dot R_B = R_B\,[\bm{\omega}^B]_{\times},
\end{gathered}
\label{eq:dynamics_eom}
\end{equation}
where $\bm q_d \in \mathbb{R}^{7}$ is the body position states and joint angles, $\bm \omega^B \in \mathbb{R}^{3}$ is the body angular velocity in body frame, and $\bm R_B \in \mathbb{R}^{3 \times 3}$ is the rotation matrix which defines the transformation from body frame to the inertial frame, i.e., $\bm x = \bm R_B \bm x^B$, where the superscript $B$ represents a vector defined in body frame. The inputs to this system are defined as follows: $\bm u_s(\bm q_k, \dot{\bm q}_k, \bm q_d, \dot{\bm q}_d)$ is the massed system spring and damping forces, and $\bm u_{a,k}(\dot{\bm q}_d, \bm \omega^B)$ is the aerodynamic force acting on the blade element $k \in [1,N_s]$. Finally, $B_s$ and $B_{a,k}$ are the mapping of forces into the generalized coordinates using principle of virtual work.

The aerodynamic forces are modeled following a simple quasi-steady blade element model for its simplicity using a similar method to our previously developed simulation model \cite{sihite_enforcing_2020}. This model follows the drag and lift coefficients developed by Dickinson \cite{sane_quasi-steady_2002}. The Aerobat's morphing wing can be categorized into four wing segments: humerus and radius wing segment on each left and right wing. Each segment is modeled as a simple rectangular wing to simplify the derivations.

\begin{figure}[t]
    \centering
    \includegraphics[width=0.6\linewidth]{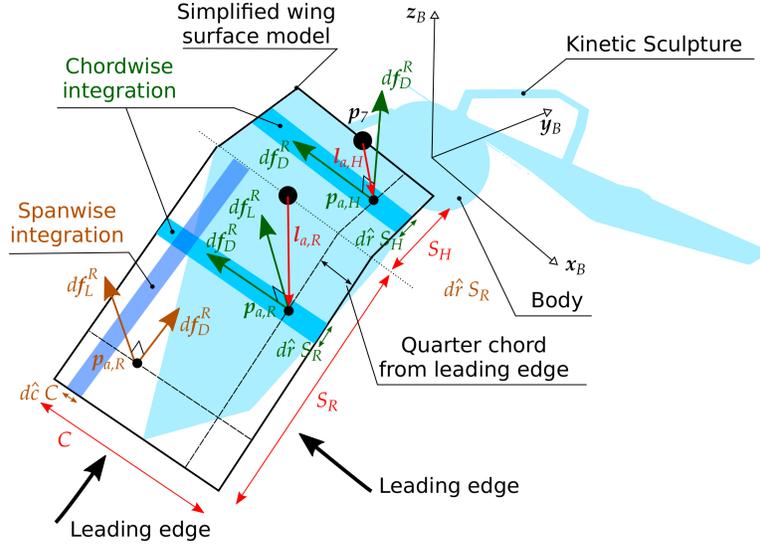}
    \caption{Illustrates the application of aerodynamic strip theory. The overall aerodynamic force is calculated both on spanwise and chordwise strip elements. Then, they are integrated to obtain the resultant aerodynamic force.}
    \label{fig:aerodynamics}
\end{figure}

Fig.~\ref{fig:aerodynamics} illustrates the blade element $k$ of the distal wing segment. Let $\bm p_{a,k}$ be the blade element $k$ aerodynamic center of pressure and $\bm p_{m,k}$ be the mid-chord point. Here, $\bm p_{a,k}$ is defined as the quarter chord distance away from the leading edge. Additionally, both $\bm p_{a,k}$ and $\bm p_{m,k}$ are defined in inertial frame. Then we can map the forces acting on $\bm p_{a,k}$ by solving for the virtual displacement $B_{a,k} = \partial \bm p_{a,k}/\partial \bm v_d$, where $\bm v_d = [\dot{\bm{ q}}_d^\top, \bm \omega^{B\,\top}]^\top$ is the system state velocity. Then the blade element's inertial aerodynamic force ($\bm u_{a,k}$) is defined as follows:
\begin{equation}
\begin{aligned}
    L_k &= \tfrac{1}{2} \rho v_{r,k}^2 \, C_L(\alpha_k) \, c_k \, \Delta s_k\\
    D_k &= \tfrac{1}{2} \rho v_{r,k}^2 \, C_D(\alpha_k) \, c_k \, \Delta s_k \\
    \bm u_{a,k} &= R_B \, R_k \, [L_k, 0, D_k]^\top,
\end{aligned}
\label{eq:strip_aero_force}
\end{equation}
where $\rho$ is the air density, $c_k$, $\Delta s_k$, and $\alpha_k$ are chord length, span width, and angle of attack of the blade element $k$, respectively. $\bm R_k$ is the rotation matrix which transform the wing segment's axis to the body frame. $v_{r,k}$ in \eqref{eq:strip_aero_force} is defined as the effective air speed at the mid-chord point $\bm p_{m,k}$ perpendicular to the wing span.

\subsection{Controller Design and Optimization}

The wing mechanism is controlled by adjusting the flapping rate and the length of the FDCs, which are the components of the massless subsystem, through the input $\bm u_k$ in \eqref{eq:kinematic_eom}. In this view, kinetic sculptures deliver dynamic morphing capabilities which are the key features in bats flight apparatus while FDCs take supervisory roles to stabilize the flight dynamics. 

Let $\dot{\theta}_1$ be the speed of the motor driven crank gear while $\bm{l} = [l_{3b}, l_{3c}, l_{8b}, l_{10b}]^\top$ be the vector containing the length of FDCs. The flapping frequency and the FDC lengths can be adjusted by a simple PD controller as shown below
\begin{equation}
\begin{aligned}
    u_g &= K_{d1} ( \omega_{ref} - \dot{\theta}_1 ) \\
    \bm{u}_p &= K_{p2} \left(\bm l_{ref} - \bm{l} \right) - K_{d2} \, \dot{\bm{l}},
\end{aligned}
\label{eq:controller}    
\end{equation}
where $K_{pi}$ and $K_{di}$ are the control gains, $\omega_{ref}$ is the desired flapping frequency, $\bm{u}_p = [u_{3b}, u_{3c}, u_{8b}, u_{10b}]^\top$, and $\bm l_{ref}$ is the desired FDC length vector. The flapping rate is set to be a constant value of 10 Hz which is the approximate flapping frequency of the Egyptian fruit bat (\textit{rousettus aegyptiacus}) which is the basis of our robot's design. The initial FDC lengths value of $\bm{l}_{ref,zp} = [7.8, 10.5, 6.2, 7.2]$ mm is found using optimization framework which give us a near stable zero path flight where the pitch angle steadily increases throughout the simulation. 

\begin{figure}[t]
    \centering
    \includegraphics[width=0.5\linewidth]{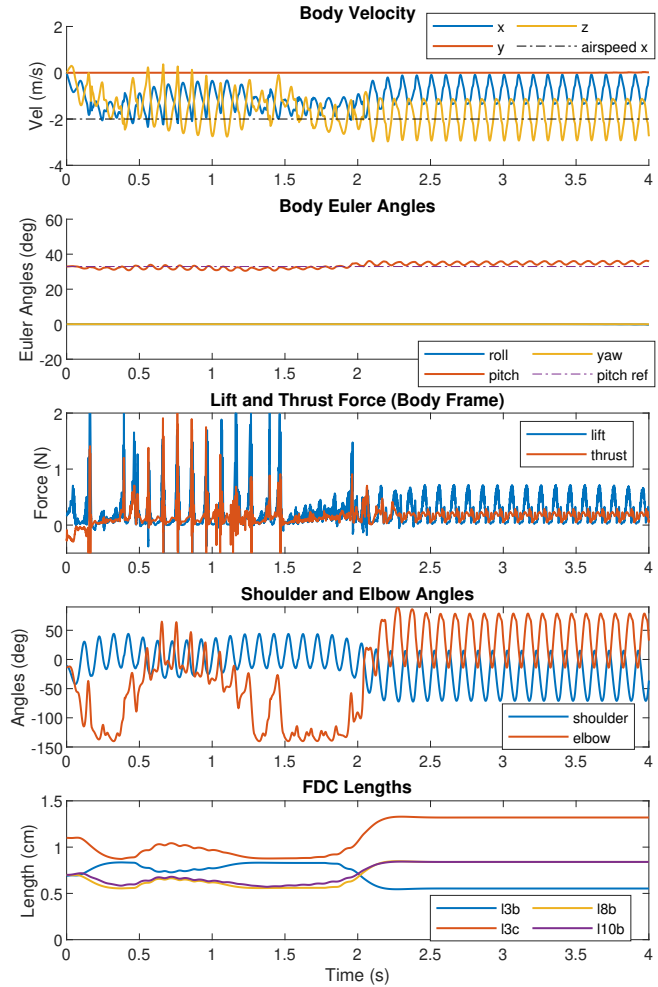}
    \caption{The simulation result using the pitch control algorithm where the controller has successfully stabilize the pitch angle and reaching a limit cycle after a transient period of approximately 2 seconds.}
    \label{fig:simulation_pitch}
\end{figure}

A controller for stabilizing the pitch can then be implemented using the following pitch stabilization controller
\begin{equation}
\begin{aligned}
    \bm l_{ref} = \bm l_{ref,zp} + K_c \, (\theta_{y,ref} - \theta_y),
\end{aligned}
\label{eq:controller_pitch}    
\end{equation}
where $\theta_{y,ref}$ is the pitch angle reference, and $K_c$ is the controller gain matrix. The gain for the controller in \eqref{eq:controller_pitch} can be found using the following optimization
\begin{equation}
    \begin{aligned}
        \min_{K_c} & \quad \textstyle{ J = \sum^N_{j=1} (w_1 \, \bm{\Pi}_j^\top\,\bm{\Pi}_j + w_2 \, \bm v_{B,j}^\top\,\bm v_{B,j} }  + w_3 \, (\theta_{y,ref} - \theta_{y,j})^2)\, \Delta t \\
        \mathrm{s. t.}  & \quad K_{c,min} \leq K_{c} \leq K_{c,max},
    \end{aligned}
\label{eq:optimization_pitch}
\end{equation}
where $\bm \Pi$ is the sum of the robot's angular momentum, $\theta_y$ is the pitch angle, and $\bm v_{B}$ is the body velocity. The subscript $i$ represents the numerical simulation time step. In this optimization, the FDC lengths are constrained using the saturation $\bm{l}_{min} \leq \bm{l}_{ref} \leq \bm{l}_{max}$ to prevent the lengths from going unbounded. The optimization found an optimal controller gain of $K_c = [0.42, -0.26, -0.38,$ $-0.097]^{\top}$ and the simulation result using this gain is shown in Fig. \ref{fig:simulation_pitch}. The robot underwent a transient period until $t = 2$s where it reaches a steady limit cycle. The pitch is stable near the target pitch angle of $33^{\circ}$ throughout the simulation, showing that this controller has successfully achieved pitch stabilization by utilizing the FDCs.

\section{CONCLUSIONS AND FUTURE WORK}

It is near impossible to copy bat's high-dimensional flight apparatus using classical feedback design paradigms based on sensing, computing, and actuation. In this work, we offered a solution towards biomimicry of bat flapping wing aerial locomotion with the control design framework incorporating morphological intelligence. Within this framework, we leveraged the armwing computational structure to adjust the robot's flapping gait through a small change in morphology using FDCs. Therefore, we subsumed part of the responsibility of closed-loop feedback under mechanical intelligence. We have extended our previous work by considering FDCs in the KS which possess supervisory roles and have shown successful stabilization of the Aerobat's longitudinal dynamics.




 

\bibliography{references-eric,references-copied} 
\bibliographystyle{spiebib} 

\end{document}